\documentclass[letterpaper, 10 pt, conference]{ieeeconf}  
\IEEEoverridecommandlockouts                              

\usepackage{soul}
\usepackage{cite}
\usepackage{graphicx,epstopdf}
\usepackage{url}
\usepackage{stfloats}
\usepackage[tight]{subfigure}
\usepackage{amsmath,bm}
\usepackage{gensymb}
\usepackage{amssymb}
\usepackage{amsfonts}   
\usepackage{amsopn}     
\usepackage{psfrag}
\usepackage{mdwtab}     
\usepackage{enumerate}
\usepackage{mdwlist}
\usepackage{accents}
\usepackage{enumerate}
\usepackage{mdwtab}     
\usepackage{mdwlist}
\usepackage{color}
\usepackage{algorithmicx}
\usepackage{algorithm} 
\usepackage{algpseudocode} 
\usepackage{acronym}
\usepackage{paralist}
\usepackage[english]{babel}
\usepackage[utf8]{inputenc}
\usepackage{hyperref}
\usepackage{optidef}

\acrodef{COM}[\textsc{COM}]{Center of Mass}
\acrodef{COP}[\textsc{COP}]{Center of Pressure}
\acrodef{DOF}[\textsc{DOF}]{Degrees of Freedom}
\acrodef{RRT}[\textsc{RRT}]{Rapidly-exploring Random Tree}
\acrodef{LIP}[\textsc{LIP}]{Linear Inverted Pendulum}
\acrodef{CBF}[\textsc{CBF}]{Control Barrier Function}
\acrodef{HZD}[\textsc{HZD}]{Hybrid Zero Dynamics}
\acrodef{MPC}[\textsc{MPC}]{Model Predictive Control}
\acrodef{MPCC}[\textsc{MPCC}]{Model Predictive Contouring Control}
\acrodef{ZMP}[\textsc{ZMP}]{Zero Moment Point}
\acrodef{DoF}[\textsc{DoF}]{Degrees of Freedom}
\acrodef{DCBF}[\textsc{DCBF}]{Discrete Control Barrier Function}
\acrodef{OSC}[\textsc{OSC}]{Operational Space Control}
\acrodef{QP}[\textsc{QP}]{Quadratic Program}
\acrodef{QCQP}[\textsc{QCQP}]{Quadratically Constrained Quadratic Program}
\acrodef{CLF}[\textsc{CLF}]{Control Lyapunov Function}
\acrodef{CLF-QP}[\textsc{CLF-QP}]{Control Lyapunov Function based Quadratic Program}
\acrodef{SLIP}[\textsc{SLIP}]{Spring Loaded Inverted Pendulum}
\acrodef{WBC}[\textsc{WBC}]{Whole-Body Controller}
\acrodef{CWC}[\textsc{CWC}]{Contact Wrench Cone}
\acrodef{SC}[\textsc{SC}]{Safe Corridor}
\acrodef{SWC}[\textsc{SWC}]{\emph{Safe Walking Corridors}}
\acrodef{IRIS}[\textsc{IRIS}]{Iterative Regional Inflation by Semidefinite}
\acrodef{UAV}[\textsc{UAV}]{Unmanned Aerial Vehicle}

\title{\LARGE \bf Overtaking Moving Obstacles with Digit: Path Following for Bipedal Robots via Model Predictive Contouring Control}

\author{Kunal S. Narkhede, Dhruv A. Thanki, Abhijeet M. Kulkarni and Ioannis Poulakakis
\thanks{K.S Narkhede, D. A. Thanki, A. M. Kulkarni, and I. Poulakakis are affiliated with the Department of Mechanical Engineering, University of Delaware, Newark, Delaware 19716, USA. {\tt\small \{kunalnk, thankid, amkulk and poulakas\}@udel.edu}}
\thanks{This work was supported in part by NSF Award MRI-2018905.}
}

\begin{document}

\maketitle
\thispagestyle{empty}
\pagestyle{empty}

\begin{abstract}
Humanoid robots are expected to navigate in changing environments and perform a variety of tasks. Frequently, these tasks require the robot to make decisions online regarding the speed and precision of following a reference path. For example, a robot may want to decide to temporarily deviate from its path to overtake a slowly moving obstacle that shares the same path and is ahead. In this case, path following performance is compromised in favor of fast path traversal. Available global trajectory tracking approaches typically assume a given---specified in advance---time parametrization of the path and seek to minimize the norm of the Cartesian error. As a result, when the robot should be where on the path is fixed and temporary deviations from the path are strongly discouraged. Given a global path, this paper presents a Model Predictive Contouring Control (MPCC) approach to selecting footsteps that maximize path traversal while simultaneously allowing the robot to decide between faithful versus fast path following. The method is evaluated in high-fidelity simulations of the bipedal robot Digit in terms of tracking performance of curved paths under disturbances and is also applied to the case where Digit overtakes a moving obstacle.  
\end{abstract}

\section{Introduction}
\label{sec: intro}
Consider a scenario in which a humanoid robot and a human share the same path with the human being in front of the robot but moving at a slower speed. The robot faces a decision between staying on the path---thus reducing its speed to avoid collisions with the human---or temporarily deviating from it to overtake the human. Global\footnote{The term ``global'' is used to refer to a path expressed with respect to an inertial frame of reference.} path following approaches for humanoid robots cannot be used directly in such situations, because they typically require a pre-specified feasible time parametrization of the reference path, thus preventing the robot from deciding \emph{online} when to be where on the path. Furthermore, these methods aim at maximizing path following accuracy, and hence do not permit temporary deviations that would result in faster path traversal. This paper describes a \ac{MPC} approach with online time reparametrization of the reference path that enables bipedal robots to balance between maximizing path following accuracy versus fast path traversal.

A wide variety of methods have been proposed for planning motions with bipedal robots. Some methods require a collection of desired footsteps, e.g.~\cite{Tedrake2015ICHR}, while others use reference velocity commands specified by a user~\cite{Barbera2013IJRR, Ijspeert2014RSS, Kheddar2019ICRA,Tsagarakis2019IROS} or by a kinematic model~\cite{Oriolo2020TRO} and translate this information to \ac{COM} target trajectories using reduced-order models like the \ac{LIP}~\cite{Kajita2001IROS}. Other methods take advantage of general sampling-based planning algorithms~\cite{Choset2005_RobotMotion} to obtain global reference paths that are then approximately followed by the robot~\cite{Agrawal2017RSS, Ghaffari2021Arxiv, Narkhede2022RAL}. Obstacle-free paths in cluttered environments can also be constructed by switching among dynamic movement primitives with limit cycle equilibrium solutions~\cite{Mohamad2016CDC, Veer2017IROS, Narkhede2023arxiv}, providing assured locomotion stability~\cite{Veer2020TAC}. In the present paper, we turn our attention on path following, that is, on global position control, assuming that a sufficiently smooth reference global path is available a priori.

In the relevant literature, parametrizing a reference global path with respect to time is typically treated in isolation from the corresponding tracking problem. For example,~\cite{Gu2018ACC, Gu2019ACC} provide a method for tracking a straight-line path under the assumption that a monotonically increasing, continuously differentiable time parametrization of the path is available \emph{a priori}. Similarly,~\cite{Xiong2021ICRA} assumes that a feasible time parametrization of the reference path is available in advance, and generates footsteps compatible with the resulting trajectory in a \ac{MPC} fashion using a hybrid extension of the classical \ac{LIP}~\cite{Kajita2001IROS}. However, decoupling the time parametrization of the path from the tracking problem may lead to inconsistencies between the resulting trajectory and the dynamics of the system. Furthermore, not allowing the robot to decide online when it should be where on the path may limit its ability to respond to changes in its surroundings.  
 
Some methods do not assume a fixed time parametrization of the reference path. For instance~\cite{Agrawal2017RSS} proposes a \ac{QCQP} to modify the heading angle for a bipedal robot model by ``shaping'' a periodic walking gait computed offline so that a reference global path can be followed; however, in this work path traversal is primarily determined by the underlying periodic trajectory. Another approach proposed in \cite{Sugihara2017ICRA} assumes a velocity profile along a reference path and converts it into walking patterns in the form of \ac{COM} and \ac{ZMP} trajectories; the velocity profile can be modified, but the ability to do this online is not explored. Translating a reference global path into a feasible \ac{COM} trajectory is studied in~\cite{Sreenath2023height}, where part of the path is converted online into a trajectory by solving a direct collocation optimization problem using the~\ac{SLIP} model.

In this work, we assume that a path is given, and we adopt the approach in~\cite{Faulwasser2009CDC} and treat the path parameter as a state variable evolving according to a dynamical system. The input to this system is determined online via a \ac{MPC} program and effectively decides how far into the path the robot goes in each step. This idea is at the core of \ac{MPCC}~\cite{Lam2010CDC}, which combines online path reparametrization and tracking within a single optimization framework, the objective of which is to minimize the error to the reference path. The key aspect of the \ac{MPCC} formulation is that the path following error is decomposed in \emph{contouring} and \emph{lag} components, allowing the system to assign different weights to maximizing tracking performance versus fast path traversal. This is different from the majority of the aforementioned methods~\cite{Agrawal2017RSS, Gu2018ACC, Gu2019ACC, Xiong2021ICRA, Sreenath2023height, Narkhede2022RAL}, which minimize the norm of the Cartesian error from the path, thus strongly penalizing deviations from it. \ac{MPCC} has been implemented initially for tracing different contours in CNC machines~\cite{Tang2012TMech}, and more recently for path following and obstacle avoidance in wheeled robots~\cite{mora2019RAL} and for realizing aggressive flight maneuvers in quadrotors~\cite{Gao2020MPCC}.

The goal of this paper is to enable bipedal robots to make decisions online regarding the speed and precision of following a reference global path. We propose a \ac{MPCC} scheme that relies on a variant of the 3D \ac{LIP} for footstep generation for bipedal robots. In contrast to previous work, our method does not require a fixed path parametrization in advance; the \ac{MPCC} just needs a reference path, determines online a feasible (discrete) time parametrization, and suggests consistent footsteps and \ac{COM} motions to the robot's low-level controller. The latter is formulated as a \ac{QP} translating these suggestions to motor torques applied to the robot. The effectiveness of the method is evaluated in simulations with a high-fidelity model of the bipedal robot Digit\footnote{https://www.agilityrobotics.com/robots\#digit}. We discuss path following performance under randomly applied external disturbances, and demonstrate the application of the method to the case where Digit overtakes an obstacle moving along its path.

\section{Path Following with Online Parametrization}
\label{sec: Background}

We consider the problem of following a global path in the robot's workspace. We describe in this section a method that allows for \emph{online} time reparametrization of the path and discuss the selection of the performance index in a \ac{MPC} program to enable the robot to decide between fast path traversal and path following performance.

\subsection{Path Following MPC with Online Reparametrization}

Consider a bipedal robot walking on flat ground. To keep the discussion general, assume that the step-to-step evolution of the robot can be predicted by a discrete-time system
\begin{equation} \label{eq:discrete_system}
    \mathrm{x}_{k+1} = f(\mathrm{x}_k,\mathrm{u}_k)
\end{equation}
where $\mathrm{x}_k \in \mathbb{R}^n$ and $\mathrm{u}_k \in \mathbb{R}^m$ are the state and input values of the system. Since we are interested in path following on the horizontal plane, we define the output 
\begin{equation}\label{eq:discrete_system_output}
    \mathrm{y}_k = h(\mathrm{x}_k) 
\end{equation}
where $\mathrm{y}_k = [x_k~y_k]^\mathsf{T} \in \mathbb{R}^2$ is the location occupied by the robot at step $k$. In Section~\ref{sec:Digit} below, we will adopt a modified \ac{LIP} as a prediction model \eqref{eq:discrete_system}-\eqref{eq:discrete_system_output} for the step-to-step dynamics of Digit.

We assume that a reference global path for the robot's \ac{COM} locations on the horizontal plane is available and that it can be represented as a sufficiently smooth curve
\begin{equation}\label{eq:path}
    \mathrm{y}^{\rm d}(\vartheta) = {\begin{bmatrix} x^{\rm d}(\vartheta) & y^{\rm d}(\vartheta) \end{bmatrix}}^{\mathsf{T}}
\end{equation}
where the parameter $\vartheta \in \Theta = [0 , \vartheta_T]$ 
measures ``progress'' along the path. Let
\begin{equation}\label{eq:theta_d}
    \phi^{\rm d}(\vartheta) = \left. \arctan 
    \left( \frac{\partial{y^{\rm d} / \partial{\vartheta}}}{\partial{x^{\rm d} / \partial{\vartheta}}} \right) \right|_{\vartheta}
\end{equation}
be the slope of the path. 

One way to discretize the path is to determine a fixed relationship between the step number $k$ and the value of $\vartheta$. This effectively corresponds to defining a trajectory $\mathrm{y}^{\rm d}(\vartheta_k) = \mathrm{y}^{\rm d}_k$ and the problem is equivalent to trajectory tracking: where the robot should be and when are specified. Instead, to allow the robot to decide \emph{online} where on the path it should go at each step, we follow~\cite{Faulwasser2009CDC} and treat the path parameter $\vartheta$ as a state evolving according to the discrete dynamical system
\begin{equation} \label{eq:mpcc_update}
   \vartheta_{k+1} = \vartheta_k + \mathrm{v}_k
\end{equation}
in which $\mathrm{v}_k$ takes values in the interval $\mathcal{V}=[0, \mathrm{v}_{\max}]$ and controls the progress of the robot along the path. In other words, the purpose of the update law \eqref{eq:mpcc_update} and its input $\mathrm{v}_k$ is to allow for reparametrizing the path online, effectively deciding how far into the path the robot travels at each step.

Given a path \eqref{eq:path}, the objective of the control system is to minimize the error between the current location $\mathrm{y}_k = h(\mathrm{x}_k)$ of the robot and the desired one specified by the parametrization $\mathrm{y}^{\mathrm{d}}(\vartheta_k)$ of the path; that is, to minimize 
\begin{equation}\nonumber
    \| e(\mathrm{x}_k, \vartheta_k) \|^2_Q = (h(\mathrm{x}_k) - \mathrm{y}^{\rm d}(\vartheta_k))^\mathsf{T}
    Q (h(\mathrm{x}_k) - \mathrm{y}^{\rm d}(\vartheta_k))
\end{equation}
where $Q = Q^\mathsf{T}>0$ is a weighting matrix. In addition to minimizing the error, we wish to penalize input effort while maximizing progress along the path. These objectives can be captured by the running cost  
\begin{equation}\label{eq:running_cost}
    J_1(\mathrm{x}_k, \vartheta_k, \mathrm{u}_k, \mathrm{v}_k) =
    \| e(\mathrm{x}_k, \vartheta_k) \|^2_Q  + \|\mathrm{u}_k\|^2_{R} - \rho |\mathrm{v}_k|^2
\end{equation}
where $R= R^\mathsf{T}>0$ and $\rho>0$; the negative sign in the last term is to maximize how far into the path the robot goes at each step. Finally, the terminal cost is defined by
\begin{equation}\label{eq:terminal_cost}
    J_2(\mathrm{x}_{k+N}, \vartheta_{k+N}) = 
    \| e(\mathrm{x}_{k+N}, \vartheta_{k+N}) \|^2_W
\end{equation}
where $W = W^\mathsf{T}>0$ is a weighting matrix.

With these definitions, path following can be realized by formulating a \ac{MPC} over a finite horizon $N > 0$ as follows:

\vspace{-0.1in}
\begin{subequations}\label{eq:MPCC_path_tracking}
\begin{IEEEeqnarray}{s'rCl}
\nonumber 
$\underset{X, U, V}{\text{minimize}}$ & \IEEEeqnarraymulticol{3}{l} {\displaystyle \sum^{k+N-1}_{\ell=k} J_1(\mathrm{x}_\ell, \vartheta_\ell, \mathrm{u}_\ell, \mathrm{v}_\ell) + J_2(\mathrm{x}_{k+N}, \vartheta_{k+N}) } 
\nonumber
\\
\nonumber
subject to & \mathrm{x}_k &=& \mathrm{x}_\mathrm{init}, ~\vartheta_k = \vartheta_\mathrm{init}
\\
\nonumber
& \mathrm{x}_{\ell+1} &\!=\!& f(\mathrm{x}_\ell,\mathrm{u}_\ell),~\vartheta_{\ell+1} = \vartheta_\ell \!+\! \mathrm{v}_\ell
\\
\nonumber
&\mathrm{x}_\ell &\!\in\!& \mathcal{X}_\ell,~\mathrm{u}_\ell \!\in\! \mathcal{U}_\ell,~\vartheta_\ell \!\in\! \Theta,~\mathrm{v}_\ell \!\in\! \mathcal{V} 
\\
\nonumber
& \mathrm{x}_{k+N} &\in& \mathcal{X}_{k+N}, ~\vartheta_{k+N} \in \Theta 
\end{IEEEeqnarray}
\end{subequations}
in which $\ell=k,..., k+N-1$ and $X = \{ \mathrm{x}_{k+1},...,\mathrm{x}_{k+N} \}$, $U = \{\mathrm{u}_k,...,\mathrm{u}_{k+N-1} \}$, $V = \{\mathrm{v}_k,...,\mathrm{v}_{k+N-1} \}$. The sets $\mathcal{X}_k \subset \mathbb{R}^n$ and $\mathcal{U}_k \subset \mathbb{R}^m$ include permissible state\footnote{If there are obstacles in the output space and $\mathcal{Y}^\mathrm{free}_k$ is the free part at step $k$, the state constraint $h(\mathrm{x}_k) \in \mathcal{Y}^\mathrm{free}_k$ determines in part the set $\mathcal{X}_k$.} and input values at step $k$, respectively.  
Given the reference path \eqref{eq:path} and initial state $\mathrm{x}_\mathrm{init}$ and path parameter $\vartheta_\mathrm{init}$, the \ac{MPC} is solved at step $k$ to provide an optimal sequence of inputs $U^*$. The first element $\mathrm{u}^*_k$ in this sequence is then applied at the robot and the process is repeated until the path is consumed.

\subsection{Cartesian and Contouring Error Formulations}

\begin{figure}[t]
\vspace{+0.1in}
    \centering
    \includegraphics[width=0.7\columnwidth]{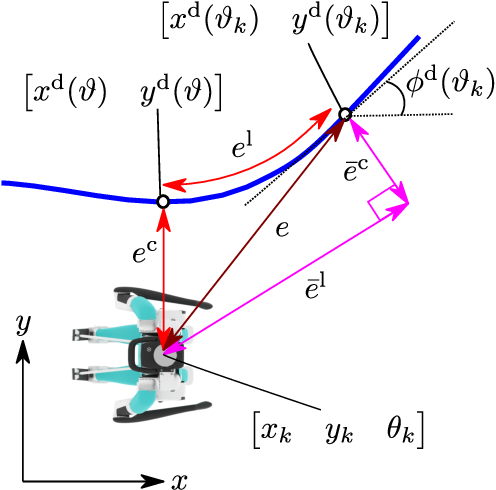}
    \vskip -10pt
    \caption{Cartesian error $e$, and contouring $e^{\rm c}$ and lag $e^{\rm l}$ error components with their corresponding approximations $\bar{e}^{\rm c}$ and $\bar{e}^{\rm l}$.
    }
    \label{fig: contouring_error}
    \vskip -15pt
    \end{figure}

To motivate our discussion, note that a common way to select the weighting matrices $Q$ and $W$ in \eqref{eq:running_cost} and \eqref{eq:terminal_cost} is 
\begin{equation}\label{eq:QP_cartesian}
    Q = \alpha I ~~\text{and}~~ W = \beta I
\end{equation}
where $\alpha>0$, $\beta>0$ and $I$ is the identity matrix. This choice corresponds to minimizing the norm of the Cartesian error; see Fig.~\ref{fig: contouring_error}. However, it does not expose the tradeoff between accurately following the path and faithfully realizing the online---not a priori specified---time parametrization \eqref{eq:mpcc_update}. As we will see in Section~\ref{sec: Results}, making this tradeoff explicit is important in certain tasks, like overtaking moving obstacles.

To capture the interplay between the spatial and temporal aspects of tracking, we decompose the error as in~\cite{Lam2010CDC} to \emph{contouring} $e^{\rm c}$ and \emph{lag} $e^{\rm l}$ components; see Fig~\ref{fig: contouring_error}. The contouring component captures the normal deviation from the path while the lag component measures the (path) distance that the robot lags behind the desired location $\mathrm{y}^\mathrm{d}(\vartheta_k)$. Since analytical expressions for $e^{\rm c}$ and $e^{\rm l}$ are not always available for general curved paths, these errors are approximated by $\bar{e}^{\rm c}$ and $\bar{e}^{\rm l}$, respectively, (see Fig.~\ref{fig: contouring_error}), which are defined as
\begin{equation}\nonumber
    \bar{e}(\mathrm{x}_k, \vartheta_k) = P(\phi^{\rm d}(\vartheta_k)) (h(\mathrm{x}_k) - \mathrm{y}^{\rm d}(\vartheta_k)) 
\end{equation}
where $\bar{e} = \begin{bmatrix}\bar{e}^{\rm c} & \bar{e}^{\rm l}\end{bmatrix}^\mathsf{T}$, $\phi^{\rm d}(\vartheta_k)$ is the slope of the path at $\vartheta_k$ computed by \eqref{eq:theta_d} and
\begin{equation}\nonumber
    P(\phi^{\rm d}(\vartheta_k)) =    
    \begin{bmatrix}
    \phantom{-}\sin(\phi^{\rm d}(\vartheta_k)) & - \cos(\phi^{\rm d}(\vartheta_k)) \\ 
    -\cos(\phi^{\rm d}(\vartheta_k)) & - \sin(\phi^{\rm d}(\vartheta_k))
    \end{bmatrix} \enspace.
\end{equation}
With these definitions, the weighted running and terminal errors can be defined as 
\begin{align}
    \nonumber
    \| \bar{e}(\mathrm{x}_k, \vartheta_k) \|^2_{\bar{Q}} &= \bar{e}(\mathrm{x}_k, \vartheta_k)^\mathsf{T}
    ~\bar{Q}~\bar{e}(\mathrm{x}_k, \vartheta_k) 
    \\
    \nonumber
    \| \bar{e}(\mathrm{x}_{k+N}, \vartheta_{k+N}) \|^2_{\bar{W}} &= \bar{e}(\mathrm{x}_{k+N}, \vartheta_{k+N})^\mathsf{T}
    ~\bar{W}~\bar{e}(\mathrm{x}_{k+N}, \vartheta_{k+N})
\end{align}
respectively, which correspond to the cost functions \eqref{eq:running_cost} and \eqref{eq:terminal_cost} with weighting matrices selected according to
\begin{align}
    \label{eq:Q_contouring}
    Q &= P^\mathsf{T}(\phi^{\rm d}(\vartheta_k))~\bar{Q}~P(\phi^{\rm d}(\vartheta_k)) 
    \\
    \label{eq:W_contouring}
    W &= P^\mathsf{T}(\phi^{\rm d}(\vartheta_k))~\bar{W}~P(\phi^{\rm d}(\vartheta_k))
\end{align}
where
\begin{equation}\label{eq:QW_contouring_diag}
    \bar{Q} = \mathrm{diag}\{\alpha_1, \alpha_2\}
    ~~\text{and}~~
    \bar{W} = \mathrm{diag}\{\beta_1, \beta_2\}
\end{equation}
are diagonal matrices with positive entries.

The selection of the weighting matrices according to \eqref{eq:Q_contouring}-\eqref{eq:W_contouring} in the definition of the cost functions of the \ac{MPC} program results in the \ac{MPCC} formulation~\cite{Lam2010CDC}. In what follows, we use the term \ac{MPCC} to emphasize that the corresponding \ac{MPC} program is formulated using the contouring/lag error decomposition. The advantage of splitting the error into contouring and lag components and using the weighting matrices \eqref{eq:Q_contouring} and \eqref{eq:W_contouring} in \eqref{eq:running_cost} and \eqref{eq:terminal_cost} is that they provide explicit control over the normal deviation of the robot from the path. By choosing the entries in \eqref{eq:QW_contouring_diag} one can prioritize between following the path accurately and keeping up with the underlying time parametrization. For example, in the case where Digit and a moving obstacle follow the same path with the obstacle being ahead of the robot, selecting $\alpha_1=\alpha_2=\alpha$ and $\beta_1=\beta_2=\beta$ results in the weighting matrices \eqref{eq:QP_cartesian} causing the robot to limit its speed in order to stay on path while avoiding collisions. In this case, the obstacle sets the pace of the robot. On the other hand, selecting suitable values  $\alpha_1 < \alpha_2$, and $\beta_1 < \beta_2$ causes the robot to prioritize fast traversal over path following, allowing it to temporarily deviate from the path to overtake the obstacle.

\section{Implementation on Digit}
\label{sec:Digit}

This section describes the 3D \ac{LIP} model that will replace \eqref{eq:discrete_system}-\eqref{eq:discrete_system_output} as a prediction model in the \ac{MPC} and discusses a \ac{QP} controller for following the \ac{MPC} suggestions. 

\subsection{Path Tracking MPC with LIP-based Predictions}

\subsubsection{3D-LIP Model}
\begin{figure}[t]
    \vspace{+0.1in}
    \centering
    \includegraphics[width=0.95\columnwidth]{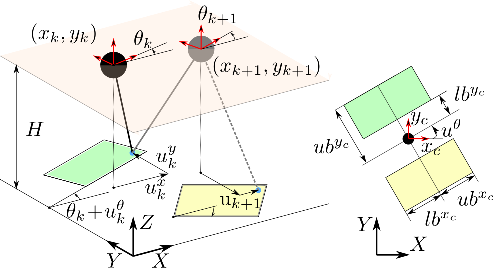}
    \caption{3D \ac{LIP} model with rechability rectangles.}
    \label{fig:reachability_constraints}
    \vskip -10pt
\end{figure}

The \ac{LIP} consists of a massless prismatic leg, on top of which is a point mass that is constrained to move in a horizontal plane located at height $H$; see Fig.~\ref{fig:reachability_constraints}. Let $(x,y)$ be the location of the point mass in the horizontal plane with respect to an inertial frame $\{X,Y,Z\}$ and let $u^x$ and $u^y$ be the distances between the ground contact point and the model's point mass along the $X$ and $Y$ axes, respectively. To ensure that the footstep locations suggested by the \ac{LIP} do not violate the geometric constraints of Digit, we restrict the swing foot's feasible landing locations to lie within a rectangular region of orientation $\theta$ with respect to the $Z$ axis, as shown in Fig.~\ref{fig:reachability_constraints}. We assume that $\theta$ is constant within each step and is updated instantaneously at the exchange of support by a step change $u^\theta$; see also~\cite{Narkhede2022RAL}.

Assuming that each step has a constant time duration $T$, the step-by-step evolution of the \ac{LIP} is given by
\begin{equation}\label{eq:LIP}
    \mathrm{x}_{k+1} = A\mathrm{x}_k + B\mathrm{u}_k
\end{equation}
where $\mathrm{x}_k \!=\! \begin{bmatrix} x_k & \dot{x}_k & y_k & \dot{y}_k & \theta_k \end{bmatrix}^\mathsf{T}$ and $\mathrm{u}_k \!=\! \begin{bmatrix} u^x_k & u^y_k & u^\theta_k \end{bmatrix}^\mathsf{T}$ 
are the state and input values at the beginning of the $k$-th step; the matrices $A$ and $B$ can be found in~\cite{Narkhede2022RAL}. To plan motions in the horizontal plane, we define
\begin{equation}\label{eq:LIP_output}
    \mathrm{y}_k = C \mathrm{x}_k
\end{equation}
where $C$ is the $2 \times 5$ matrix so that $C \mathrm{x} = \begin{bmatrix} x & y \end{bmatrix}^\mathsf{T}$. 

\subsubsection{Constraints}
To ensure that the motions suggested by the \ac{LIP} are within the capabilities of Digit, we introduce the following constraints. The first constraint captures the reachability rectangles for foot placement and can be described by
\begin{equation}\label{eq:rectangles}
    \mathrm{lb}_k \leq \mathsf{Rot}(\theta_k + u^\theta_k)^\mathsf{T} \mathrm{u}_k \leq \mathrm{ub}_k
\end{equation}
where $\mathrm{lb}_k = \begin{bmatrix} lb^{x_c}_k & lb^{y_c}_k & lb^{\theta}_k\end{bmatrix}^\mathsf{T}$ (resp. $\mathrm{ub}_k$) includes lower (resp. upper) bounds determining the rectangles, as shown in Fig.~\ref{fig:reachability_constraints} and $\mathsf{Rot}(\theta_k + u^\theta_k)$ denotes the 3D rotation matrix about the $Z$ axis by the angle $\theta_k + u^\theta_k$. 

To avoid infeasible motions, we introduce a constraint that limits the distance covered by the \ac{LIP} during each step within the interval $[\delta_{\min}, \delta_{\max}]$ with suitably chosen $0 \leq \delta_{\min}< \delta_{\max}$. Formally, we require    
\begin{equation}\label{eq:travelled_distance}
    \delta_\mathrm{min} \leq \sqrt{\Delta x_k^2 + \Delta y_k^2} \leq \delta_\mathrm{max}
\end{equation}
where $\Delta x_k = x_{k+1} - x_k$ and  $\Delta y_k = y_{k+1} - y_k$. Note here that when the underlying path has small curvature, \eqref{eq:travelled_distance} has a similar effect with the constraint $\mathrm{v}_k \in [0, \mathrm{v}_{\max}]$ in \eqref{eq:mpcc_update}; however, for paths with considerable curvature, the two constraints influence the system in different ways: \eqref{eq:mpcc_update} limits how far into the path the robot can travel in each step while \eqref{eq:travelled_distance} constraints the Cartesian distance travelled per step.

Finally, obstacles---either static or moving---can be incorporated via additional constraints in the \ac{MPC} as in~\cite{Narkhede2022RAL}. For simplicity, let us consider a circular obstacle of radius $r$, the location $\mathrm{p}^\mathrm{obs}_k$ of which is known at each step $k$. Collisions can then be avoided by imposing the barrier constraint
\begin{equation}\label{eq:CBF_constraint}
    b\left( A \mathrm{x}_k + B \mathrm{u}_k \right) \geq (1-\gamma) b\left(\mathrm{x}_{k}\right)
\end{equation}
where the parameter $\gamma \in (0,1)$ and $b(\mathrm{x}_k) = (C \mathrm{x}_k - \mathrm{p}^\mathrm{obs}_{k})^{\mathsf{T}}  
(C\mathrm{x}_k - \mathrm{p}^\mathrm{obs}_{k}) - r^2$. This procedure can be extended to avoid collisions with multiple obstacles and can include obstacles of elliptical shape as well; see~\cite{Narkhede2022RAL} for more details.

\subsubsection{MPC formulation}
The path tracking \ac{MPC} program of Section~\ref{sec: Background} can be adapted as follows:
\begin{subequations}\label{eq:MPCC}
\begin{IEEEeqnarray}{s'rCl'rCl'rCl}
\nonumber 
$\underset{X, U, V}{\text{minimize}}$ & \IEEEeqnarraymulticol{9}{l} {\displaystyle \sum^{k+N-1}_{\ell=k} J_1(\mathrm{x}_\ell, \vartheta_\ell, \mathrm{u}_\ell, \mathrm{v}_\ell) + J_2(\mathrm{x}_{k+N}, \vartheta_{k+N}) } 
\\
\nonumber 
subject to & \mathrm{x}_k &=& \mathrm{x}_{\rm init},~\vartheta_k = \vartheta_{\rm init} 
\\
\nonumber
& \mathrm{x}_{\ell+1} &=& A\mathrm{x}_\ell+B\mathrm{u}_\ell,~\vartheta_{\ell+1} = \vartheta_{\ell} + \mathrm{v}_{\ell} 
\\
\nonumber 
& (\mathrm{x}_{\ell}, \mathrm{u}_\ell) &\in& \mathcal{XU}_\ell ,~\vartheta_\ell \in \Theta, ~\mathrm{v}_\ell \in \mathcal{V} 
\\
\nonumber 
&\vartheta_{k+N} &\in& \Theta
\end{IEEEeqnarray}
\end{subequations}
where $\ell=k,..., k+N-1$, the cost functions $J_1$, $J_2$ are given by \eqref{eq:running_cost}, \eqref{eq:terminal_cost} with the output $h(\mathrm{x})$ replaced by \eqref{eq:LIP_output}, and
\begin{align}
    \nonumber
    \mathcal{XU}_{\ell} = \big\{ (\mathrm{x}_\ell, \mathrm{u}_\ell)  ~|&~ \mathrm{lb}_\ell \leq \mathsf{Rot}(\theta_\ell + u^\theta_\ell)^\mathsf{T} \mathrm{u}_\ell \leq \mathrm{ub}_\ell, 
    \\
    \nonumber
     &~\text{and}~ \delta_\mathrm{min} \leq \sqrt{\Delta x_\ell^2 + \Delta y_\ell^2} \leq \delta_\mathrm{max}, 
    \\
    \nonumber
    &~\text{and}~ b\left( A \mathrm{x}_\ell + B \mathrm{u}_\ell \right) \geq (1-\gamma) b\left(\mathrm{x}_{\ell}\right) \big\} 
\end{align}
combines the constraints \eqref{eq:rectangles}-\eqref{eq:CBF_constraint} in a single state- and input-dependent set. The symbols $\Theta$ and $\mathcal{V}$ correspond to intervals defined in Section~\ref{sec: Background} above. At step $k$, the \ac{MPC} requires $\mathrm{x}_\mathrm{init}$ and $\vartheta_\mathrm{init}$. The former is available though feedback of the robot's state and the latter is computed by 
\begin{equation} \label{eq:init_parameter_optimization}
    \vartheta_\mathrm{init} = \underset{\vartheta}{\mathrm{argmin}} \|C \mathrm{x}_\mathrm{init}-\mathrm{y}^\mathrm{d}(\vartheta)\|^2
\end{equation}
which corresponds to the value of $\vartheta$ that minimizes the distance between the robot and the path; i.e., $\vartheta_\mathrm{init}$ is the point on the path closest to the robot's location $C \mathrm{x}_\mathrm{init}$.

\begin{figure}[b]
    \centering
    \includegraphics[width=0.95\columnwidth]{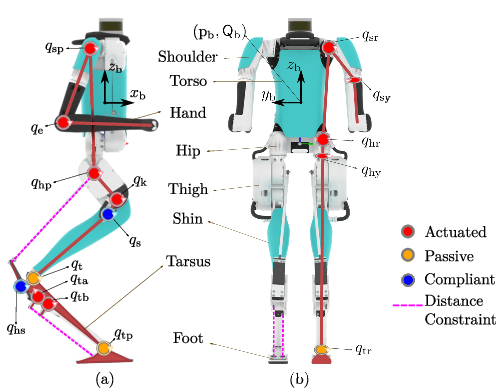}
    \caption{The bipedal robot Digit and relevant coordinates. (a) Side view; the kinematic loops actuating the tarsus and foot are highlighted with the dotted magenta lines. (b) Back view.}
    \label{fig:DigitStructureandFoot}
\end{figure}

\subsection{Closing the loop with Digit}
\label{subsec: Digit Implementation}

The bipedal robot Digit shown in Fig.~\ref{fig:DigitStructureandFoot} has a total of 30 \ac{DOF}s; 20 are directly actuated, 4 correspond to leaf springs at the legs and are passive, and 6 represent the position and orientation of a body-fixed frame relative to an inertial frame. The robot with a suitable choice of coordinates is shown in Fig.~\ref{fig:DigitStructureandFoot}.
\begin{figure*}[t!]
    \vspace{+0.1in}
    \centering
    \includegraphics[width=\textwidth]{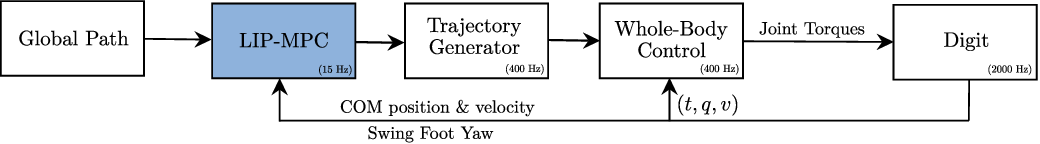}
    \caption{Implementation of the MPC with Cartesian or contouring error representations on Digit.}
    \label{fig:ImplementationOnDigit}
    \vskip -10pt
\end{figure*}

The structure of Digit's controller is shown in Fig.~\ref{fig:ImplementationOnDigit}. With feedback of Digit's \ac{COM} location, velocity and swing foot position and orientation, the initial state $\mathrm{x}_\mathrm{init}$ required by the \ac{MPC} is specified. The \ac{MPC} is then solved as described above to suggest a target footstep location and orientation, and \ac{COM} position for the next step. The output of the \ac{MPC} is then fed to the trajectory generator shown in Fig.~\ref{fig:ImplementationOnDigit}, which uses interpolation to obtain the \ac{COM} and swing foot position and orientation trajectories for Digit; see~\cite{Narkhede2022arXiv} for relevant details. These trajectories are then used to define the outputs
\begin{equation} \nonumber\label{eq:WBC_Obj}
    \begin{bmatrix}
    \left(\ac{COM}~\mathrm{Position}\right)_{3 \times 1} \\
    \left(\mathrm{Torso~Orientation}\right)_{3 \times 1} \\
    \left(\mathrm{Swing~Foot~Position}\right)_{3 \times 1} \\
    \left(\mathrm{Swing~Foot~Orientation}\right)_{3 \times 1} \\
    \end{bmatrix}_{12 \times 1} \enspace.
\end{equation}
The objective of the \ac{QP} controller is then to minimize the weighted sum of the square norm of 
\begin{itemize}
    \item the error from the desired outputs computed above;
    \item the error from a desired constant arm configuration; and
    \item Digit's centroidal angular momentum; 
\end{itemize}
while satisfying physical constraints such as torque saturation and contact wrench constraints. Due to space limitations, we do not provide further details on the formulation of the \ac{QP} here; these can be found in~\cite{Narkhede2022arXiv}. 

\begin{figure*}[b!]
    \vskip -15pt
    \centering
    \subfigure[]{\includegraphics[width=0.215\textwidth]{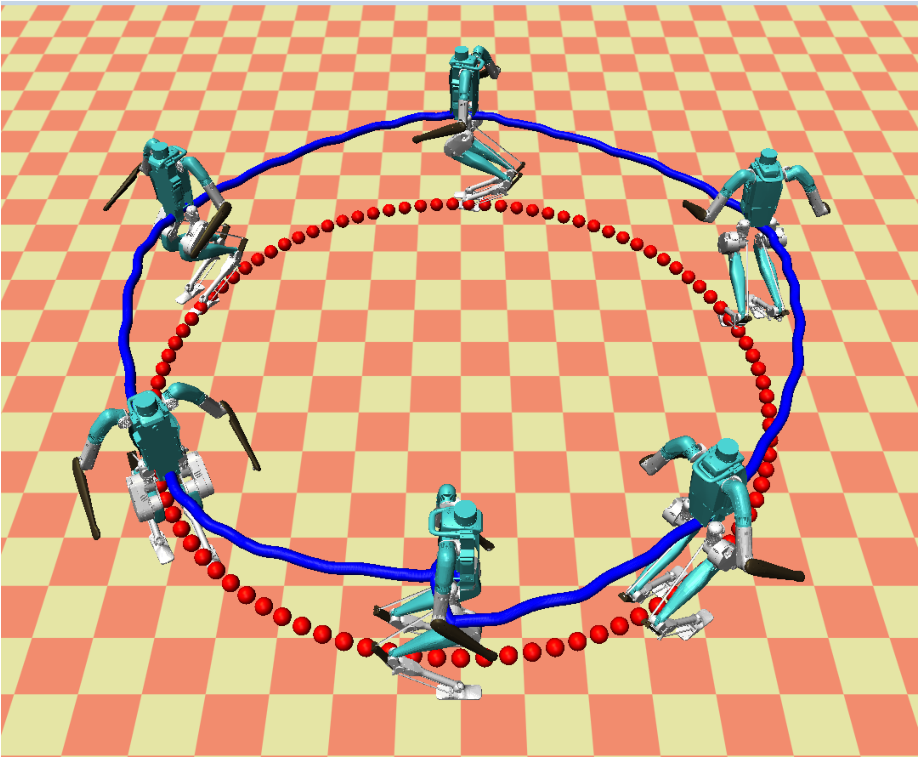} \label{fig: st_line}}
    \subfigure[]{\includegraphics[width=0.23\textwidth]{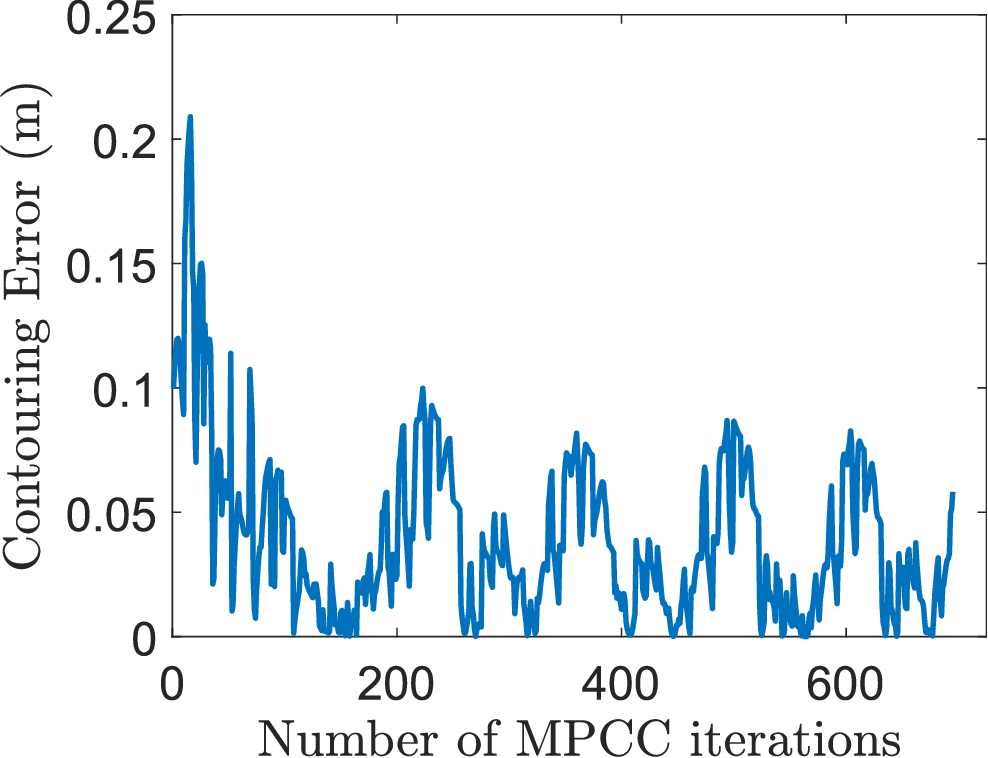} \label{fig: st_cont_error}}
    \subfigure[]{\includegraphics[width=0.215\textwidth]{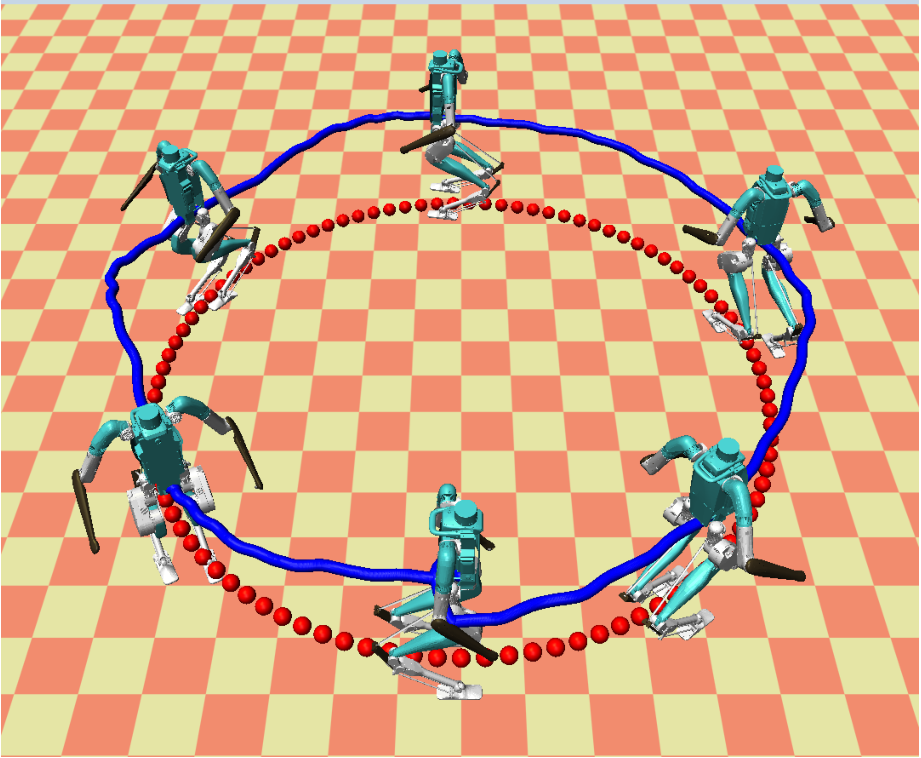} \label{fig: st_line_dist}}
    \subfigure[]{\includegraphics[width=0.23\textwidth]{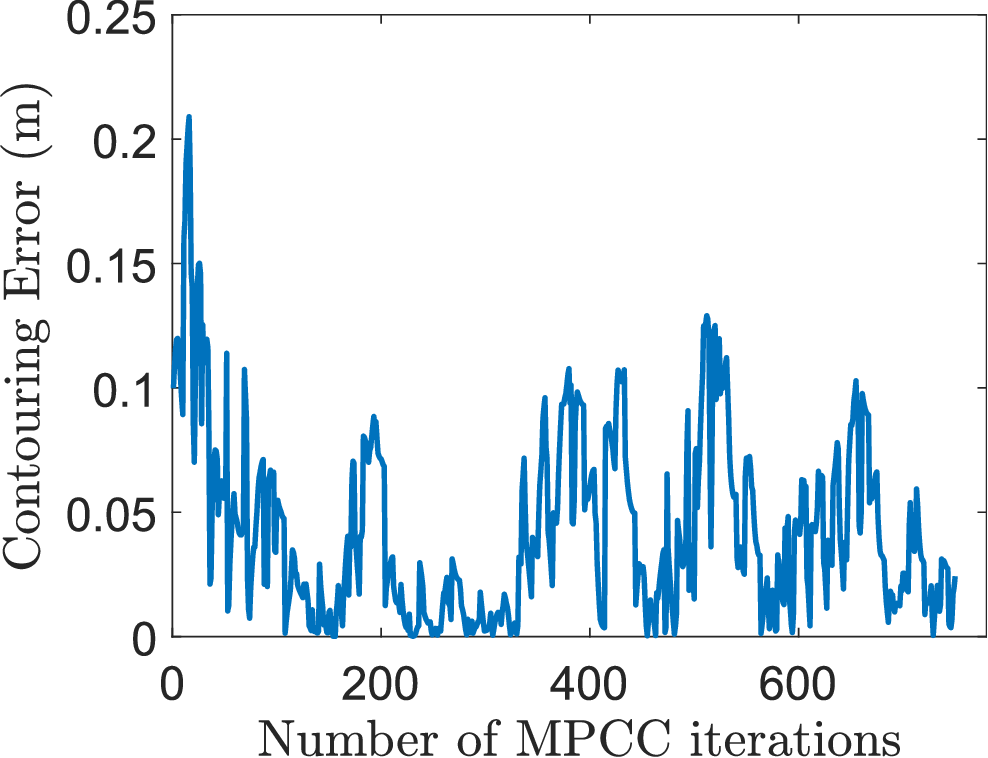} \label{fig: st_cont_error_dist}}
    \vskip -10pt
    \caption{Tracking performance of the \ac{MPCC}; the reference path is in red and the robot's \ac{COM} path is in blue. (a) Digit follows a circular path with no external disturbances. (b) Contouring error for the case of Fig.~\ref{fig: st_line}. (c) Digit follows the same path as Fig.~\ref{fig: st_line} under randomly applied external disturbances. (d) Corresponding  contouring error.}
    \label{fig: tracking_performance}
    \vskip -10pt
\end{figure*}

\begin{figure*}[t!]
    \vskip +10pt
    \centering
    \subfigure[]{\includegraphics[]{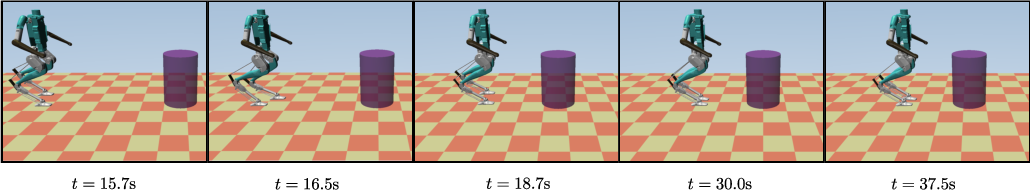} \label{fig:trail_snapshot}}
    \vskip -5pt
    \subfigure[]{\includegraphics[]{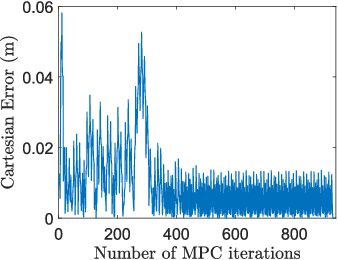} \label{fig:cartesian_error}}
    \subfigure[]{\includegraphics[]{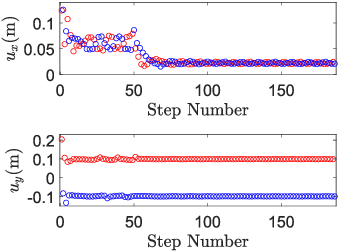} \label{fig:trail_input}}
    \subfigure[]{\includegraphics[]{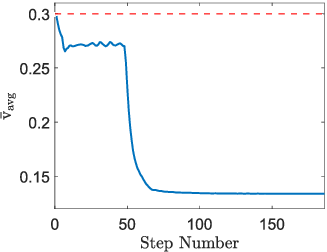} \label{fig:trail_parameter}}
    \subfigure[]{\includegraphics[]{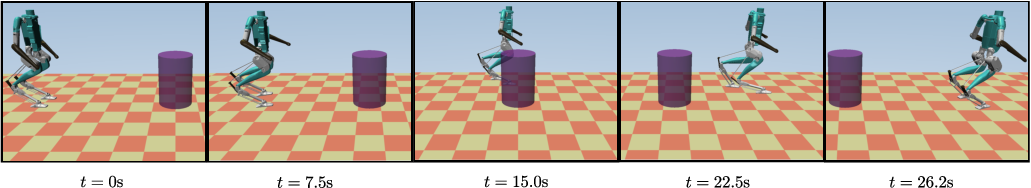} \label{fig:overtaking_snapshot}}
    \vskip -5pt
    \subfigure[]{\includegraphics[]{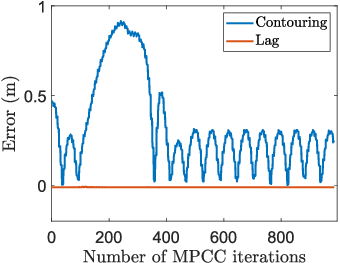} \label{fig:contouring_error}}
    \subfigure[]{\includegraphics[]{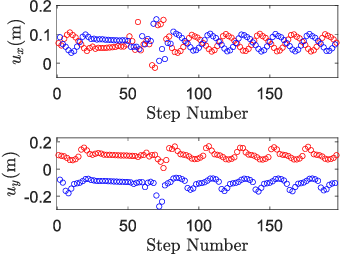} \label{fig:overtaking_input}}
    \subfigure[]{\includegraphics[]{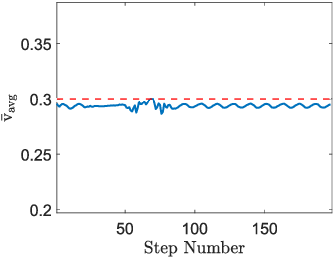} \label{fig:overtaking_parameter}}
    \caption{Cartesian versus contouring error formulations. (a) Digit trailing behind a moving obstacle via Cartesian error \ac{MPC}. (b) Cartesian error norm. (c) Right (blue) and left (red) footsteps. (d) Average path update parameter (blue) and max value (red). (e) Digit overtaking the moving obstacle via \ac{MPCC}. (f) Contouring (blue) and lag (red) errors. (g) Right (blue) and left (red) footsteps. (h) Average path update parameter (blue) and max value (red).}
    \label{fig: obstacle_avoidance}
    \vskip -15pt
\end{figure*}

\section{Results}
\label{sec: Results}

This section evaluates the method using high-fidelity simulations of the bipedal robot Digit tasked with following a reference path. All simulations are performed using the physics-based simulator MuJoCo~\cite{Todorov2012IROS_mujoco}, with the simulation loop operating at $2\mathrm{kHz}$. The \ac{MPC} is programmed in Python using Casadi~\cite{Andersson2019_CasADi} and solved at $15\mathrm{Hz}$ using interior point solver IPOPT~\cite{Wchter2006IPOPT}. The average computation time for solving the \ac{MPC} including the optimization \eqref{eq:init_parameter_optimization} for $\vartheta_{\rm init}$ is $12 \mathrm{ms}$. The low-level \ac{QP} is formulated using CVXPY~\cite{diamond2016cvxpy} and solved at $400\mathrm{Hz}$ with its pre-shipped solver. All computations were carried out on an Intel PC equipped with $\mathrm{i}7$ processor ($2.60$ GHz) and $16\mathrm{GB}$ RAM. The input weights $R = \mathrm{diag}\{100,100,5\}$, the upper bound on the parameter update $\mathrm{v}_{\rm max} = 0.3$ and the horizon $N=5$ are kept fixed for all cases in this section. Refer to~\cite{videolink} for simulation video.

\subsection{Path Following Performance under Disturbances}

We discuss here the performance of the \ac{MPCC} controller in following curved paths under disturbances; we include results only for a circular path with radius $2\mathrm{m}$, noting that other curved paths can also be followed with similar performance. Here, the weight matrices for the contouring and lag errors are selected as $\bar{Q} = \bar{W} = \mathrm{diag}\{300,3\}$ and $\rho = 10$. Figure~\ref{fig: st_line} shows the path following performance of the \ac{MPC} controller without external disturbances and Fig.~\ref{fig: st_cont_error} shows the corresponding contouring error, indicating satisfactory tracking performance. On the other hand, Fig.~\ref{fig: st_line_dist} shows tracking of the same path under random external disturbances. The disturbances are forces with magnitude sampled from a uniform distribution over the interval $[-50,50]~\mathrm{N}$ and are applied at Digit's torso over a duration of $0.1\mathrm{s}$ at random instants separated by at most $2\mathrm{s}$. As seen from Fig.~\ref{fig: tracking_performance}, the \ac{MPCC} is able to track the given path even in presence of disturbances with occasional spikes appearing when the disturbances are present.

\subsection{Overtaking Moving Obstacles: A Case Study}

We consider here a scenario in which Digit is tasked with following a given path that is shared by a moving obstacle ahead of the robot. We compare two cases: In the first, the objective of the \ac{MPC} is to minimize the norm of the Cartesian error from the path, while in the second, to minimize the contouring and lag errors corresponding to the \ac{MPCC} formulation. The goal of this comparison is to demonstrate the benefits of the \ac{MPCC}, which allows the robot to decide between emphasizing path following quality versus fast overall path traversal. For simplicity, and without loss of generality, we consider a straight-line path and a cylindrical obstacle with radius $r_{\rm obs} = 0.25\mathrm{m}$ moving at constant speed along the path. 
\vspace{-0.1in}

In the first case, we assume that the obstacle is moving with a slow speed $0.1\mathrm{m/s}$, and we select equal gain matrices $Q = W = 200 I$ and $\rho = 50$ in \eqref{eq:running_cost}-\eqref{eq:terminal_cost}, corresponding to the Cartesian error norm. As we can see from Fig.~\ref{fig:trail_snapshot}, Digit quickly covers the distance from the obstacle, but cannot pass it because the objective of minimizing the Cartesian error norm does not allow temporary deviations from the path. This is evident from Fig.~\ref{fig:cartesian_error}, which shows that the norm of the Cartesian error remains below $0.06\mathrm{m}$. To keep following the path with high accuracy while avoiding collisions with the obstacle, Digit adjusts its stepping online to take shorter steps; this can be seen in Fig.~\ref{fig:trail_input}, where the step size $u^x$ along the $X$ direction decreases when the obstacle constraint is activated. As a measure of how ``deep'' into the path the robot moves, Fig.~\ref{fig:trail_parameter} presents the average path update parameter $\mathrm{v}_{\rm avg} = (1/N) \sum^{N-1}_{\ell=0} \mathrm{v}_\ell$ over the horizon $N$ for each iteration of the \ac{MPC}. The obstacle constraint prevents the \ac{MPC} from suggesting large updates in the path parameter. 

For the \ac{MPCC}, the obstacle is moving at a higher speed $0.6\mathrm{m/s}$, comparable with Digit's speed. Here, the weight matrices are $\bar{Q} = \bar{W} = \mathrm{diag}\{1,1000\}$ and $\rho = 50$. This choice emphasizes the lag component of the cost over the contouring component; thus, the controller is more tolerant to deviations from the path, prioritizing fast overall path traversal. It can be seen from Fig.~\ref{fig:overtaking_snapshot} that Digit overtakes the moving obstacle. This is achieved because, as shown in Fig.~\ref{fig:contouring_error}, the controller allows the robot to temporarily deviate from the path---the contouring error increases---while the lag error is kept almost zero. It is clear from Fig. ~\ref{fig:overtaking_input} that the stepping size of the robot increases in the $X$ direction, while small deviations in the $Y$ direction allow Digit to skirt around the obstacle and eventually leave it behind. This behavior is due to the larger weight chosen for the lag component of the error, as compared to the contouring component. From Fig.~\ref{fig:overtaking_parameter}, we can see that the \ac{MPCC} constantly keeps the parameter $\mathrm{v}_{\rm avg}$ close to the maximum bound $\mathrm{v}_{\rm max}$, thus encouraging the robot to go as far into the path as the \ac{MPCC} constraints allow.

 \section{Conclusion}
 \label{sec: Conclusion}

This paper presented a \ac{MPCC} scheme for foot placement that enables bipedal robots to make online decisions on the speed and accuracy with which they follow a planned global path. Unlike other approaches, the proposed method does not require specifying a feasible trajectory or a speed profile in advance. Using physics-based simulations with the bipedal robot Digit, we discussed aspects of \ac{MPCC} that allow temporary deviations from the reference path to favor fast traversal of the overall path. We demonstrated these capabilities in the case where Digit overtakes a moving obstacle that shares its path. Our future work will focus on experimentally validating the proposed approach and on integrating \ac{MPCC} with safe walking corridors~\cite{Narkhede2022RAL} to ensure fast path traversal and safety in highly-cluttered spaces.

\bibliographystyle{IEEEtran}
\bibliography{MPCC_arxiv_ref}

\begin{thebibliography}{10}
\providecommand{\url}[1]{#1}
\csname url@samestyle\endcsname
\providecommand{\newblock}{\relax}
\providecommand{\bibinfo}[2]{#2}
\providecommand{\BIBentrySTDinterwordspacing}{\spaceskip=0pt\relax}
\providecommand{\BIBentryALTinterwordstretchfactor}{4}
\providecommand{\BIBentryALTinterwordspacing}{\spaceskip=\fontdimen2\font plus
\BIBentryALTinterwordstretchfactor\fontdimen3\font minus
  \fontdimen4\font\relax}
\providecommand{\BIBforeignlanguage}[2]{{%
\expandafter\ifx\csname l@#1\endcsname\relax
\typeout{** WARNING: IEEEtran.bst: No hyphenation pattern has been}%
\typeout{** loaded for the language `#1'. Using the pattern for}%
\typeout{** the default language instead.}%
\else
\language=\csname l@#1\endcsname
\fi
#2}}
\providecommand{\BIBdecl}{\relax}
\BIBdecl

\bibitem{Tedrake2015ICHR}
R.~Tedrake, S.~Kuindersma, R.~Deits, and K.~Miura, ``A closed-form solution for
  real-time {ZMP} gait generation and feedback stabilization,'' in \emph{Proc.
  IEEE Int. Conf. Humanoid Robots}, 2015, pp. 936--940.

\bibitem{Barbera2013IJRR}
J.~Alcaraz-Jiménez, D.~Herrero-Pérez, and H.~Martínez-Barberá, ``Robust
  feedback control of {ZMP}-based gait for the humanoid robot {Nao},''
  \emph{Int. J. Robot. Res.}, vol.~32, no. 9-10, pp. 1074--1088, 2013.

\bibitem{Ijspeert2014RSS}
S.~Faraji, S.~Pouya, and A.~Ijspeert, ``Robust and agile {3D} biped walking
  with steering capability using a footstep predictive approach,'' in
  \emph{Proc. Robot. Sci. Syst.}, 2014.

\bibitem{Kheddar2019ICRA}
A.~Tanguy, D.~De~Simone, A.~I. Comport, G.~Oriolo, and A.~Kheddar,
  ``Closed-loop {MPC} with dense visual {SLAM} - stability through reactive
  stepping,'' in \emph{Proc. IEEE Int. Conf. Robot. Autom.}, 2019, pp.
  1397--1403.

\bibitem{Tsagarakis2019IROS}
S.~Xin, R.~Orsolino, and N.~Tsagarakis, ``Online relative footstep optimization
  for legged robots dynamic walking using discrete-time model predictive
  control,'' in \emph{Proc. IEEE/RSJ Int. Conf. Intell. Robots Syst.}, 2019,
  pp. 513--520.

\bibitem{Oriolo2020TRO}
N.~Scianca, D.~De~Simone, L.~Lanari, and G.~Oriolo, ``{MPC} for humanoid gait
  generation: {Stability} and feasibility,'' \emph{IEEE Trans. Robot.},
  vol.~36, no.~4, pp. 1171--1188, 2020.

\bibitem{Kajita2001IROS}
S.~Kajita, F.~Kanehiro, K.~Kaneko, K.~Yokoi, and H.~Hirukawa, ``The {3D} linear
  inverted pendulum mode: a simple modeling for a biped walking pattern
  generation,'' in \emph{Proc. IEEE/RSJ Int. Conf. Intell. Robots Syst.}, 2001,
  pp. 239--246.

\bibitem{Choset2005_RobotMotion}
H.~Choset, K.~Lynch, S.~Hutchinson, G.~Kantor, W.~Burgard, L.~E. Kavraki, and
  S.~Thrun, \emph{Principles of Robot Motion: Theory, Algorithms, and
  Implementation}.\hskip 1em plus 0.5em minus 0.4em\relax MIT Press, 2005.

\bibitem{Agrawal2017RSS}
A.~Agrawal and K.~Sreenath, ``Discrete control barrier functions for
  safety-critical control of discrete systems with application to bipedal robot
  navigation,'' in \emph{Proc. Robot. Sci. Syst.}, 2017.

\bibitem{Ghaffari2021Arxiv}
S.~Teng, Y.~Gong, J.~W. Grizzle, and M.~Ghaffari, ``Toward safety-aware
  informative motion planning for legged robots,'' \emph{arXiv:2103.14252},
  2021.

\bibitem{Narkhede2022RAL}
K.~S. Narkhede, A.~M. Kulkarni, D.~A. Thanki, and I.~Poulakakis, ``A sequential
  {MPC} approach to reactive planning for bipedal robots using safe corridors
  in highly cluttered environments,'' \emph{IEEE Robot. Automat. Lett.},
  vol.~7, no.~4, pp. 11\,831--11\,838, 2022.

\bibitem{Mohamad2016CDC}
M.~S. {Motahar}, S.~{Veer}, and I.~{Poulakakis}, ``Composing limit cycles for
  motion planning of {3D} bipedal walkers,'' in \emph{Proc. IEEE Conf. Decis.
  Control}, 2016, pp. 6368--6374.

\bibitem{Veer2017IROS}
S.~{Veer}, M.~S. {Motahar}, and I.~{Poulakakis}, ``Almost driftless navigation
  of {3D} limit-cycle walking bipeds,'' in \emph{Proc. IEEE/RSJ Int. Conf.
  Intell. Robots Syst.}, 2017, pp. 5025--5030.

\bibitem{Narkhede2023arxiv}
K.~S. Narkhede, M.~S. Motahar, S.~Veer, and I.~Poulakakis, ``Reactive gait
  composition with stability: Dynamic walking amidst static and moving
  obstacles,'' \emph{arXiv:2303.16165}, 2023.

\bibitem{Veer2020TAC}
S.~Veer and I.~Poulakakis, ``Switched systems with multiple equilibria under
  disturbances: Boundedness and practical stability,'' \emph{IEEE Trans. Autom.
  Control}, vol.~65, no.~6, pp. 2371--2386, 2020.

\bibitem{Gu2018ACC}
Y.~Gu, B.~Yao, and C.~S. George~Lee, ``Straight-line contouring control of
  fully actuated 3-{D} bipedal robotic walking,'' in \emph{Proc. Amer. Control
  Conf.}, 2018, pp. 2108--2113.

\bibitem{Gu2019ACC}
Y.~Gao and Y.~Gu, ``Global-position tracking control of a fully actuated {NAO}
  bipedal walking robot,'' in \emph{Proc. Amer. Control Conf.}, 2019, pp.
  4596--4601.

\bibitem{Xiong2021ICRA}
X.~Xiong, J.~Reher, and A.~D. Ames, ``Global position control on underactuated
  bipedal robots: Step-to-step dynamics approximation for step planning,'' in
  \emph{Proc. IEEE Int. Conf. Robot. Autom.}, 2021, pp. 2825--2831.

\bibitem{Sugihara2017ICRA}
H.~Atsuta, H.~Nozaki, and T.~Sugihara, ``Smooth-path-tracking control of a
  biped robot at variable speed based on dynamics morphing,'' in \emph{Proc.
  IEEE Int. Conf. Robot. Autom.}, 2017, pp. 4116--4121.

\bibitem{Sreenath2023height}
Z.~Li, J.~Zeng, S.~Chen, and K.~Sreenath, ``Autonomous navigation of
  underactuated bipedal robots in height-constrained environments,'' \emph{Int.
  J. Robot. Res.}, 2023, early access, doi: 10.1177/02783649231187670.

\bibitem{Faulwasser2009CDC}
T.~Faulwasser, B.~Kern, and R.~Findeisen, ``Model predictive path-following for
  constrained nonlinear systems,'' in \emph{Proc. IEEE Conf. Decis. Control},
  2009, pp. 8642--8647.

\bibitem{Lam2010CDC}
D.~Lam, C.~Manzie, and M.~Good, ``Model predictive contouring control,'' in
  \emph{Proc. IEEE Conf. Decis. Control}, 2010, pp. 6137--6142.

\bibitem{Tang2012TMech}
L.~Tang and R.~G. Landers, ``Predictive contour control with adaptive feed
  rate,'' \emph{IEEE/ASME Trans. Mechatronics}, vol.~17, no.~4, pp. 669--679,
  2012.

\bibitem{mora2019RAL}
B.~{Brito}, B.~{Floor}, L.~{Ferranti}, and J.~{Alonso-Mora}, ``Model predictive
  contouring control for collision avoidance in unstructured dynamic
  environments,'' \emph{IEEE Robot. Automat. Lett.}, vol.~4, no.~4, pp.
  4459--4466, 2019.

\bibitem{Gao2020MPCC}
J.~Ji, X.~Zhou, C.~Xu, and F.~Gao, ``{CMPCC}: Corridor-based model predictive
  contouring control for aggressive drone flight,'' in \emph{Proc. Int. Symp.
  Exp. Robot.}, 2020, p. 37–46.

\bibitem{Narkhede2022arXiv}
K.~S. Narkhede, A.~M. Kulkarni, D.~A. Thanki, and I.~Poulakakis, ``A sequential
  {MPC} approach to reactive planning for bipedal robots,''
  \emph{arXiv:2205.00156}, 2022.

\bibitem{Todorov2012IROS_mujoco}
E.~Todorov, T.~Erez, and Y.~Tassa, ``{M}u{J}o{C}o: A physics engine for
  model-based control,'' in \emph{Proc. IEEE/RSJ Int. Conf. Intell. Robots
  Syst.}, 2012, pp. 5026--5033.

\bibitem{Andersson2019_CasADi}
J.~A.~E. Andersson, J.~Gillis, G.~Horn, J.~B. Rawlings, and M.~Diehl,
  ``{CasADi}-{A} software framework for nonlinear optimization and optimal
  control,'' \emph{Math. Program. Comput.}, vol.~11, no.~1, pp. 1--36, 2019.

\bibitem{Wchter2006IPOPT}
A.~W{\"a}chter and L.~T. Biegler, ``On the implementation of an interior-point
  filter line-search algorithm for large-scale nonlinear programming,''
  \emph{Math. Program.}, vol. 106, pp. 25--57, 2006.

\bibitem{diamond2016cvxpy}
S.~Diamond and S.~Boyd, ``{CVXPY}: {A} {P}ython-embedded modeling language for
  convex optimization,'' \emph{J. Mach. Learn. Res.}, vol.~17, no.~83, pp.
  1--5, 2016.

\bibitem{videolink}
\BIBentryALTinterwordspacing
Simulation results. [Online]. Available: \url{https://youtu.be/5FBwor9ROK4}
\BIBentrySTDinterwordspacing

\end{thebibliography}

\end{document}